\documentclass{article} 

\usepackage{iclr2027_conference,times}

\usepackage{amsmath,amsfonts,bm}

\def\eqref#1{equation~\ref{#1}}

\def\1{\bm{1}}

\DeclareMathAlphabet{\mathsfit}{\encodingdefault}{\sfdefault}{m}{sl}
\SetMathAlphabet{\mathsfit}{bold}{\encodingdefault}{\sfdefault}{bx}{n}

\usepackage[noend]{algpseudocode}
\usepackage{float}
\usepackage{subcaption}
\usepackage{tabularx}
\usepackage{booktabs}
\usepackage{algorithm2e}
\usepackage{multirow}
\usepackage{booktabs}
\usepackage{tablefootnote}
\usepackage{threeparttable}
\usepackage{enumitem} 
\usepackage{tcolorbox}
\usepackage{verbatim}
\usepackage{longtable}

\usepackage{amsmath,amssymb}
\usepackage{pifont}
\usepackage{graphicx}
\usepackage{subcaption}
\usepackage[T1]{fontenc}
\usepackage{xcolor}
\usepackage{listings}
\tcbuselibrary{listingsutf8}

\usepackage{fancyhdr}
\usepackage{fvextra}
\fvset{fontsize=\footnotesize, breaklines=true}
\usepackage{tcolorbox}
\tcbuselibrary{skins} 

\definecolor{codegreen}{rgb}{0,0.6,0}
\definecolor{codegray}{rgb}{0.5,0.5,0.5}

\definecolor{backcolour}{RGB}{245,248,250}
\definecolor{emph}{RGB}{166,88,53}
\definecolor{nightblue}{RGB}{9,49,105}
\definecolor{keywords}{RGB}{207,33,46}
\definecolor{lightpurple}{RGB}{130,81,223}

\lstdefinestyle{mystyle}{
    backgroundcolor=\color{backcolour},   
    commentstyle=\color{codegreen},
    keywordstyle=\color{keywords},
    stringstyle=\color{nightblue},
    basicstyle=\ttfamily\footnotesize,
    breakatwhitespace=false,         
    breaklines=true,                 
    captionpos=b,                    
    keepspaces=true,                 
    numberstyle=\small\color{codegray},
    numbers=left,                    
    numbersep=5pt,   
    xleftmargin=0.2cm,
    aboveskip=0.2cm,
    belowskip=0.1cm,
    showspaces=false,                
    showstringspaces=false,
    showtabs=false,                  
    tabsize=2,
    frame=shadowbox,
    emph={sem_filter,sem_join,sem_topk,sem_sim_join,sem_agg,sem_map,sem_extract,sem_cluster_by,sem_search,sem_index,load_sem_index,sem_partition_by,sem_group_by},
    emphstyle={\color{lightpurple}},
}

\newif\ifcomments
\commentstrue
\ifcomments
    \providecommand{\rev}[1]{{\color{blue}{/* rev: #1 */}}}
    \providecommand{\liana}[1]{{\color{purple}{/* lp: #1 */}}}
    \providecommand{\rishabh}[1]{{\color{orange}{/* rr: #1 */}}}
    \providecommand{\claude}[1]{{\color{teal}{/* claude: #1 */}}}
    
    \providecommand{\ion}[1]{{\color{blue}{/* ion: #1 */}}}

\else
    \providecommand{\rev}[1]{}
    \providecommand{\liana}[1]{}
    \providecommand{\rishabh}[1]{}
    \providecommand{\ion}[1]{}
    \providecommand{\claude}[1]{}

\fi

\newcommand{\heading}[1] {{\noindent{\textbf{\emph{#1}}} }}

\newcommand{\name}{\textsc{TokenGate}\xspace}

\definecolor{logocolor}{RGB}{138, 72, 213}                

\newcounter{samplebox}
\renewcommand{\thesamplebox}{\arabic{samplebox}}
\newtcolorbox[use counter=samplebox]{samplebox}[2][]{
  enhanced,
  colback=white, 
  colframe=black, 
  boxrule=0.5pt, 
  arc=5pt, 
  outer arc=5pt, 
  title={Box~\thesamplebox: #2}, 
  fonttitle=\color{white},
  colbacktitle=black, 
  fontupper=\small,
  #1,
}

\usepackage{hyperref}
\usepackage{url}

\newif\ifarxiv
\arxivtrue      
\newcommand{\arxivcodeurl}{https://github.com/tokengate-models/TokenGate}

\ifarxiv
  \iclrfinalcopy
  \makeatletter
  \def\@maketitle{\vbox{\hsize\textwidth
    {\centering\LARGE\sc \@title\par}
    \begin{tabular}[t]{l}\bf\rule{\z@}{24pt}\@author\end{tabular}%
    \vskip 0.3in minus 0.1in}}
  \makeatother
  
  \newcommand{\paperauthors}{\arxivtitleblock}
  \newcommand{\aftertitle}{%
    \fancyhead{}\renewcommand{\headrulewidth}{0pt}%
    \begingroup
      \renewcommand{\thefootnote}{}%
      \footnotetext{\raggedright\arxivcorrespondence}%
    \endgroup
  }
\else
  \newcommand{\paperauthors}{}          
  \newcommand{\aftertitle}{}
\fi

\begin{document}

\title{\name: Adapting LLMs for Fast and Memory-Efficient Prefill with\\ Learned Token Pruning}


\author{\paperauthors}

\maketitle
\aftertitle

\begin{abstract}
Expert human readers process large inputs (e.g., code repositories and long documents) by skimming to closely read only what matters. On the other hand, language models process all input tokens through each model layer, resulting in significant processing costs for long-context inputs. 
We introduce \name, a small, drop-in component that extends existing trained models to enable end-to-end learnable token pruning over long-context inputs for fast and memory-efficient prefill. 
To enable end-to-end learning despite the non-differentiable top-$K$ operator for token pruning, we introduce \emph{token-gated attention}, a mechanism inspired by Mixture-of-Experts that re-weights attention logits using token importance scores.
Across diverse tasks, \name achieves $1.4$-$11.3\times$ faster time-to-first token (TTFT) and $1.5$-$3.4\times$ lower peak memory at no accuracy degradation for prefill in transformers.
Compared to the best-performing prior methods for token-pruning, \name gives up to $6.4\times$ faster prefill and $2\times$ lower peak memory.
\name natively extends both standard transformers (e.g., Qwen3) and hybrid-attention models (e.g., Qwen3.5) with simple lightweight adaptation comparable to standard parameter-efficient fine-tuning.
We make our code available at: \href{https://github.com/tokengate-models/TokenGate}{\texttt{https://github.com/tokengate-models/TokenGate}}.


\end{abstract}



\section{Introduction}
\label{sec:intro}

Long-context inputs underpin exciting recent language model deployments, where the model's ability to process vast swaths of personalized user data and live content during inference has led to transformative capabilities in deep research, large-scale document processing, coding agents, and data agents~\citep{noauthor_introducing_nodate-1, noauthor_gemini_nodate-1, noauthor_claude_nodate-1, ClaudeCode, Cursor, Copilot}.
When processing large inputs, like code or long documents, experts readers skim to closely read only what matters.
Yet, transformers process every token of their input through each layer, whether or not the token bears critical information. The cost of this full read is attention compute that grows quadratically with input length, as well as MLP compute and KV-cache memory that grow linearly, making long-context prefill a key bottleneck in model serving.

\vspace{-.1cm}
\begin{figure*}[t]
  \centering
  \includegraphics[width=\linewidth]{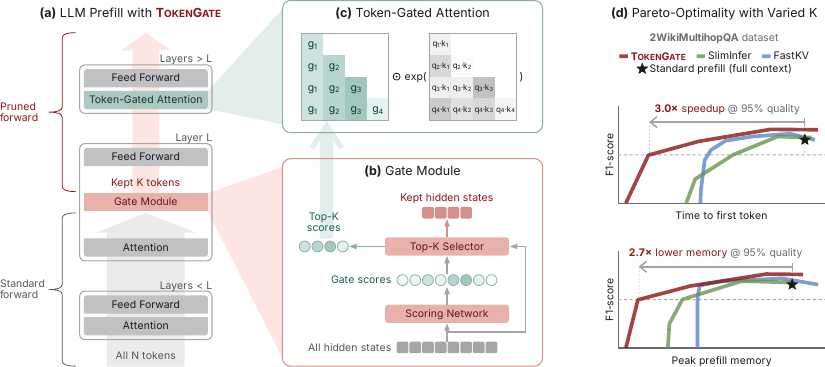}
    \vspace{-.1cm}
  \caption{\footnotesize \textbf{Overview of \name.} \textbf{(a)}~Layers up to $L$ prefill all $N$ context tokens as usual; at layer $L$ the gate module keeps $K$ of them, and every layer above $L$ processes only those $K$ tokens. \textbf{(b)}~The gate module scores each hidden state with a small scoring network, and a top-$K$ selector forwards the kept hidden states together with their importance scores. \textbf{(c)}~Token-gated attention multiplies each attention weight by the gate value $g_j$ of its key, which enables end-to-end training. \textbf{(d)}~On 2WikiMultihopQA question answering, with Qwen3-1.7B, \name{} pushes the Pareto frontier of accuracy versus time-to-first-token and peak prefill memory; at $95\%$ of the full-context accuracy it prefills $3.0\times$ faster and with $2.7\times$ less memory, outperforming the best training-free token pruning methods, like FastKV~\citep{FastKV} and SlimInfer~\citep{SlimInfer}.
  }
  \label{fig:intro}
  \vspace{-.4cm}
\end{figure*}

Recent research proposes to address this efficiency challenge for long-context prefill through two primary approaches. First, one body of work proposes \emph{training-free token pruning}, using attention-based scoring heuristics to prune token hidden states, forwarding only the highest-scoring tokens to upper layers~\citep{FastKV, SlimInfer, PromptDistill, GemFilter, PyramidInfer, LazyLLM, DASH, SpAtten}. These methods are inexpensive to deploy, requiring no model training, but demonstrate limited speedups before degrading accuracy.
Alternatively, a large body of work instead develops \emph{architectural modifications} to the standard transformer architecture~\citep{Transformer}. These methods offer promising prefill-efficiency improvements, but are expensive to train, often requiring pre-training from scratch and substantially different hardware and software than existing ecosystems~\citep{UniversalTransformer, MixtureOfRecursions, ACT, MixtureOfDepths, CoLT5, LTP, RouterTuning, Mamba, Mamba2}.
To bridge this gap, we ask the question, \emph{can we endow existing trained models with improved long-context prefill efficiency while maintaining strong accuracy?}

In this work, we introduce \name, a small, drop-in component that extends transformer-based models, enabling end-to-end trainable token pruning over long-context inputs for efficient prefill.
The model forward pass using \name decomposes into three parts, as Figure~\ref{fig:intro}a shows. First, layers below the gate module perform a standard forward pass, processing all tokens through each layer. Second, tokens pass through the gate module, which produces a score for each token, retaining only the top-$K$ for continued processing. Third, the layers above the gate module process the selected tokens with \emph{token-gated attention}, a mechanism that re-weights the attention computation using the generated importance scores. \name reduces prefill compute and memory because pruned tokens are never forwarded through the post-gate layers and these layers never generate KV cache entries.

Token pruning uses a non-differentiable top-$K$ operator, which presents a challenge for end-to-end learning. We introduce \emph{token-gated attention} to overcome this issue.
Our method takes inspiration from Mixture-of-Experts (MoE)~\citep{SparseMoE, SwitchTransformer}, which achieve sparsity over \emph{activated model parameters} by gating expert sub-networks. We use a similar form, but instead weight attention logits to achieve \emph{sparsity over processed tokens}. Using this technique, we demonstrate that we can adapt existing trained models with cost similar to standard adaptation techniques. For instance, we can fine-tune existing models with \name using simple next-token-prediction loss, with no auxiliary loss and total train time similar to supervised fine-tuning on the base model. The resulting \name model matches or exceeds accuracy over the trained base model, while significantly reducing prefill latency and memory.

We evaluate \name on diverse tasks, including code completion, multi-hop QA, multi-doc summarization, medical extraction, and needle-in-a-haystack, using both traditional transformer models and a recent hybrid attention one. Overall, we show the following findings \textbf{(1)} \name \emph{reduces TTFT and peak prefill memory} by $1.4$-$11.3\times$ and $1.5$-$3.4\times$, respectively, at no accuracy degradation compared to the finetuned model baseline with full-context, and outperforms all training-free token pruning methods, by up to $6.4\times$ faster prefill (TTFT) and $2\times$ lower peak prefill memory relative to the next-best baseline,  
\textbf{(2)} \name \emph{natively extends existing recent hybrid-attention architectures}, like Qwen3.5-2B~\citep{Qwen35}, where it achieves $3.8\times$ lower TTFT over the full-context baseline at no accuracy degradation, and $2.1\times$ lower TTFT than the next best-performing token-pruning method at fixed accuracy.
\textbf{(3)} \emph{Training models is efficient} with \name, incurring minimal overhead compared to standard parameter-efficient fine-tuning techniques, like LoRA~\citep{LoRA}.

Overall, we summarize our contributions as follows:
\begin{itemize}[leftmargin=*]
    \item We introduce \name, a small, drop-in component that extends existing trained models to enable end-to-end learnable token pruning for fast and memory-efficient prefill.
    \item We propose \emph{token-gated attention}, a mechanism for training models end-to-end despite the gate module's non-differentiable top-$K$ operator.
    \item Our systematic evaluation finds that \name reduces TTFT and peak prefill memory by up to $6.4\times$ and $2\times$, respectively, compared to the next best-performing baselines, \name natively extends transformer-based models, including recent hybrid-attention architectures.
    Further, training is efficient, with total train time comparable to standard parameter-efficient fine-tuning techniques.
\end{itemize}

\section{Related Works}
\label{sec:relwork}

\heading{Training-free Token Pruning.}
One body of prior work optimizes prefill using training-free token-pruning methods (e.g., FastKV, SlimInfer, DASH, and GemFilter)~\citep{FastKV, SlimInfer, PromptDistill, GemFilter, PyramidInfer, LazyLLM, DASH, SpAtten}. These methods prune token hidden states at a mid-layer of the model using attention-based heuristics to rank tokens, forwarding only the highest-scoring tokens to upper layers of the model. This body of work is closest to ours, however, unlike our work, these methods are not trainable. As we observe in (Section~\ref{sec:results}), these methods offer limited speedups before degrading accuracy because they employ heuristic attention-based token-ranking functions that under-perform our end-to-end trained scoring network, and they prevent co-adaptation of the model backbone with the pruning function, which our end-to-end trainable design enables.

\heading{Architectural Modifications.}
Several lines of work explore architectural modifications to the self-attention-based transformer~\citep{Transformer}, reducing the compute and memory of processing long-context inputs.
First, one body of work studies architectures that achieve \emph{adaptive compute per token} through dynamic token routing or learned pruning decisions, but are expensive to train. These works require pre-training from scratch~\citep{UniversalTransformer, MixtureOfRecursions, ACT, MixtureOfDepths, CoLT5} or expensive fine-tuning with auxiliary loss terms that fix the token budget during train time, requiring re-training for each inference-time token budget~\citep{LTP, RouterTuning, PoWERBERT, DynamicViT}. 
In contrast to all of these works, \name natively extends popular pre-trained language models and is simple and efficient to train. We show that with a single fine-tuning pass, the resulting \name model generalizes across input lengths and token budgets (Section~\ref{sec:results}, Appendix~\ref{app:ctx_scaling}).

Other works on \emph{sub-quadratic architectures} for language modeling introduce promising, new algorithms that reduce the \emph{per token} compute and memory of attention operations, including new compute primitives and sparse attention mechanisms~\citep{Mamba, Mamba2, DeltaNet, GatedDeltaNet, LinearAttention, TTT, SeerAttention, DSA, NSA, MoBA, DuoAttention, WGKV, MInference, FlexPrefill}. 
However, unlike \name, these architectures still process all tokens through each model layer.
In Section~\ref{sec:results} we demonstrate \name natively extends recent hybrid architectures with sub-quadratic layers~\citep{Qwen35}, and enables even more efficient prefill.

\heading{Conditional Computation.} Mixture-of-experts layers achieve sparsity over the activated model parameters, making it possible to efficiently scale model capacity~\citep{SparseMoE, SwitchTransformer, DeepSeekV2}. Our work is inspired by the success of this line of work, however, we focus on achieving sparsity over a different axis, the input context length, for scalable long-context prefill.

\heading{Decode Optimizations.} A large body of work targets the cost of decoding rather than prefill, evicting or merging KV-cache entries after prefill~\citep{H2O, SnapKV, StreamingLLM, FastGen, PyramidKV, KVzip}, and performing early exit or per-token routing for decoded tokens~\citep{CALM, LayerSkip}. These methods do not improve prefill efficiency and are complementary to prefill optimizations like ours.

\section{Preliminaries}
\label{sec:prelim}

\heading{Transformer Models and Notation.}
Given a sequence of tokens $x = (x_1, \dots, x_N)$, the language model embeds each token $x_i$ into a vector $h^{0}_i \in \mathbb{R}^{d_{\mathrm{model}}}$ and passes the sequence through a stack of $N_L$ transformer layers. Layer $\ell \in \{0, \dots, N_L - 1\}$ reads the \emph{residual stream} $h^{\ell}_i$ and applies a self-attention sub-layer followed by a token-wise MLP. Both sub-components are preceded by a normalization layer and add their output back to the residual stream, as given by:
\begin{align}
a^{\ell}_i \;=\; h^{\ell}_i + \mathrm{Attn}^{\ell}\big(\mathrm{LN}_1(h^{\ell})\big)_i, \qquad
h^{\ell+1}_i \;=\; a^{\ell}_i + \mathrm{MLP}^{\ell}\big(\tilde a^{\ell}_i\big), \qquad
\tilde a^{\ell}_i \;=\; \mathrm{LN}_2(a^{\ell}_i). \label{eq:layer}
\end{align}
We call $a^{\ell}_i$ the post-attention residual of token $i$ at layer $\ell$. The MLP acts on each position independently, whereas attention mixes information across positions.

In standard self-attention, the normalized inputs $\mathrm{LN}_1(h^{\ell})$ are projected to queries, keys, and values $q_i, k_j, v_j \in \mathbb{R}^{d_{\mathrm{head}}}$, and token $i$ attends causally to itself and the tokens before it:
\begin{align}
\mathrm{Attn}_i \;=\; \sum_{j \le i} \alpha_{ij}\, v_j, \quad
\alpha_{ij} \;=\; \operatorname{softmax}_{j \le i}\big(z_{ij}\big) \;=\; \frac{\exp(z_{ij})}{\sum_{j' \le i} \exp(z_{ij'})}, \quad
z_{ij} \;=\; \frac{q_i^{\top} k_j}{\sqrt{d_{\mathrm{head}}}}. \label{eq:softmax}
\end{align}

\heading{Prefill, decode, and the KV cache.}
Generation proceeds in two phases. \emph{Prefill} runs Eq.~\ref{eq:layer} over the whole input of $N$ tokens and stores every layer's keys and values, the \emph{KV cache} $\{(k^{\ell}_j, v^{\ell}_j)\}_{j \le N}$. \emph{Decode} then produces one token per step, computing attention against the cache and appending to it. The cost of both phases grows with the input length $N$. Prefill compute scales as $O(N^2 \cdot N_L)$ for attention and $O(N \cdot N_L)$ for the MLPs, and the KV cache occupies $O(N \cdot N_L)$ memory.

\begin{figure}[t]
  \vspace{-.1cm}
  \centering
  \includegraphics[width=\linewidth]
  {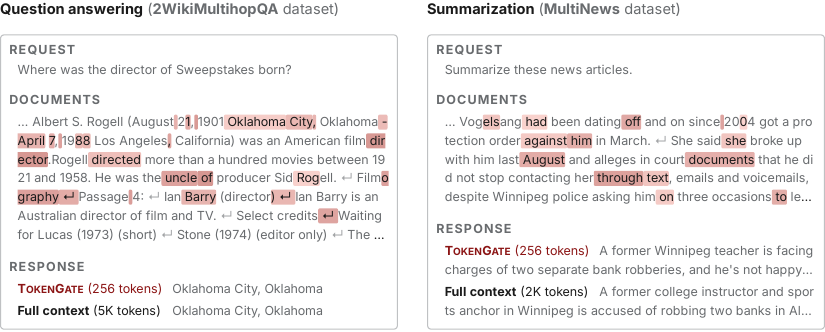}
  \caption{ \footnotesize Visualization of \name applied on sample inputs from the 2WikiMultihopQA dataset (left) and MultiNews dataset (right), with $K=256$ kept tokens, highlighted in red (darker denotes higher token importance score by the gate).
  Detailed visualizations of real samples from all datasets are in Appendix~\ref{app:kept_tokens}.
  }
  \label{fig:viz}
    \vspace{-.2cm}
\end{figure}

\subsection{Problem Setting}
\label{sec:prelim:problem}
We consider an LLM input prompt $x = [x_R, x_D]$, where $x_R = (x_1, \dots, x_r)$ is a request instruction or question and $x_D = (x_{r+1}, \dots, x_N)$ are the ``documents'' needed to respond to it, such as research articles, code repositories, medical records, legal documents, or financial filings. 
In practice, $x_D$ may be supplied by the user or retrieved by a system or an agent given $x_R$. 
As an example, a user uploads a document study as a PDF to an AI assistant and asks ``what are the side effects reported in this study?''~\citep{NotebookLM, ChatGPTFiles, ClaudeProjects, GeminiWorkspace, M365Copilot, AcrobatAI, PerplexityFiles, ChatPDF}. Here, $x_R$ corresponds to the user's question, and $x_D$ corresponds to the attached document, which can be thousands to hundreds of thousands of tokens. Figure~\ref{fig:viz} shows two such input prompts from our datasets.

In this setting, the documents are provided only at request time, so its KV cache cannot be precomputed. Our goal is two-fold: we aim to \emph{maximize quality}, measured by task accuracy, and \emph{minimize cost}, both prefill time-to-first-token (TTFT), which is the latency users experience before seeing any response, and prefill memory footprint, which limits throughput and concurrent user requests.

\section{Trainable Token Pruning with \name}
\label{sec:method}
\name is a drop-in neural network component, which places a gate module within a specified transformer layer\footnote{The design generalizes to multiple gate modules at different layers, each selecting from the tokens kept by the layers below it and contributing to the token-gated attention in layers above.}, $L$, to score and prune tokens, as shown in Figure~\ref{fig:intro}.
In this section, we describe how \name enables \emph{trainable} token pruning using two key components, the \emph{gate module} (Section~\ref{subsec:gatemodule}) and \emph{token-gated attention} (Section~\ref{subsec:tokengatedattn}). We will then overview an end-to-end training recipe for adapting existing models (Section~\ref{subsec:e2etraining}).

Overall, the full \name forward pass with gating at layer $L$ decomposes into three phases. First, tokens pass through layers below the gate module, using a \emph{standard forward} pass. Second, the \emph{gate module} processes all tokens, producing a score for each, and selecting a subset to prune. Third, the model performs a \emph{pruned forward}, processing only selected tokens in layers above the gate module, using \emph{token-gated attention}, which re-weights attention scores using the importance scores.
\name reduces both the compute and memory footprint during prefill by preventing model layers after the gate module from processing pruned tokens or generating KV-cache entries for them.

\subsection{Gate Module}
\label{subsec:gatemodule}

The gate module is placed within a transformer layer, $L$, and selects a subset of tokens to continue processing through the remaining model layers during prefill. As Figure~\ref{fig:intro}b shows, the gate module consists of two simple sub-components: a \emph{scoring network} and a \emph{top-$K$ selector}.

\textbf{Scoring Network.}
For each token $j$ in the set of prunable tokens $\mathcal{P}$, an MLP reads its normalized post-attention residual at layer $L$, $\tilde a^{L}_j$ (Eq.~\ref{eq:layer}), and emits a scalar score $\tilde b_j$, which we normalize along the sequence with RMSNorm~\citep{RMSNorm} into the
importance score $b_j$, following:
\begin{align}
    \tilde b_j = \mathrm{MLP}_\theta\big(\tilde a^{L}_j\big), \qquad b_j = \frac{\tilde b_j}{\sqrt{\frac{1}{|\mathcal{P}|} \sum_{j' \in \mathcal{P}} \tilde b_{j'}^2 + \epsilon}} \ \text{ for } j \in \mathcal{P}, \qquad b_j := 0 \ \text{ for } j \in \mathcal{N}, \label{eq:scorer}
\end{align}
where the scores are fixed to zero for the \emph{non-prunable tokens} $\mathcal{N}$,
which can consist of answer tokens during model training, decoded tokens, or any optionally specified tokens during preill, e.g., a user question.
The MLP is initialized to output zero, so that $b_j = 0$ for all $j$ at the start of training.

\textbf{Top-$K$ Selector.} The gate selects the token-set $S$, consisting of the $K$ highest-scoring prunable tokens, as well as any non-prunable tokens, for continued processing:
\begin{align}
    S = \mathrm{top}\text{-}K_{j \in \mathcal{P}}\, b_j \;\cup\; \mathcal{N}. \label{eq:select}
\end{align}

The residual stream for pruned tokens is stopped at layer $L$, and the selected tokens $S$ are forwarded through the layer $L$ MLP and layers $\ell > L$.

\subsection{Token-Gated Attention}
\label{subsec:tokengatedattn}
The gate module's top-$K$ selector introduces a non-differentiability, creating a challenge for end-to-end learning. To overcome this challenge, we introduce \emph{token-gated attention}, which integrates token importance scores generated by the gate module into upper model layers, providing a gradient path. Our method is inspired by Mixture-of-Experts (MoE)~\citep{SparseMoE}.
In MoE, a router network weights the outputs of expert sub-networks to achieve \emph{sparsity over activated model parameters}.
In contrast, token-gated attention re-weights attention logits
to achieve \emph{sparsity over processed tokens}.

Token-gated attention multiplies each token's attention weight by the gate value $g_j$, computed from it's importance score $b_j$ (Eqs.~\ref{eq:gated}), in layers $\ell > L$. Specifically, for self-attention, in contrast to Eq.~\ref{eq:softmax}, the attention weights with token-gated attention take the form of Eq.~\ref{eq:gated}.
At the start of training with a base model, token-gated attention is equivalent to top-$K$ attention
as the importance scores are $0$.
\begin{align}
    \alpha_{ij} = \frac{\exp(z_{ij})\, g_j}{\sum_{j' \in S,\, j' \le i} \exp(z_{ij'})\, g_{j'}}, \qquad i, j \in S, \qquad g_j = \operatorname{softmax}_{j \in S}( b_j) \label{eq:gated}
\end{align}

We make two modifications to this basic form: \emph{head-wise scaling}, to allow different heads to have different scales for the gate value, and \emph{temperature scaling}, to enable length generalization so that trained models perform robustly at various pruning ratios. We ablate both decisions in Appendix~\ref{app:additional_results}.

\heading{Head-wise.} We scale the normalized importance scores $b_j$ by a learnable gain factor, $\gamma^{\ell h}$, indexed by layer $\ell$ and head $h$, and the gate value is the \emph{softmax} of the result (Eq.~\ref{eq:gate}).
\begin{align}
    g^{\ell h}_j = \operatorname{softmax}_{j \in S}\big(\mathrm{softplus}(\gamma^{\ell h})\, b_j\big) \label{eq:gate}
\end{align}

Intuitively, $\gamma^{\ell h}$ learns how strongly each attention head gates the attention logits. The softplus keeps every gain positive, so a higher importance score corresponds to higher attention weight. At initialization,
$\text{softplus}({\gamma^{\ell h})}$ is set to the standard deviation of layer $\ell$'s logits
so that it is
on the same scale as the attention logits.

\heading{Temperature Scaling.} Pruning shrinks the number of keys each query sees from its original position $p_i$ to its rank $m_i$ among kept tokens. This pruning sharpens the softmax relative to the full-context model. We use temperature scaling~\citep{TemperatureScaling, PostHocReversal} to recalibrate the softmax. The temperature is learnable, positive, and scaled by the log-ratio of the two counts (Eq.~\ref{eq:temp}), inspired by SSMax~\citep{ssmax}. The parameter $\sigma$ is initialized such that temperature is $1$ at full context.
\begin{align}
    \tau^{\ell h}_i = \mathrm{softplus}(\sigma^{\ell h})\cdot \frac{\log m_i}{\log \max (2, p_i)} \label{eq:temp}
\end{align}

\heading{Putting everything together.} Substituting Eq.~\ref{eq:gate} into Eq.~\ref{eq:gated}, we can write the token-gated attention as 
\begin{align}
    \alpha^{\ell h}_{ij} = \operatorname{softmax}_{j \in S,\, j \le i} \Big(\tau^{\ell h}_i \big( \tfrac{q_i^{\top} k_j}{\sqrt{d_{\mathrm{head}}}} + \mathrm{softplus}(\gamma^{\ell h})\, b_j\big)\Big). \label{eq:tokengated}
\end{align}
This form is straightforward to implement as an efficient kernel based on FlashAttention~\citep{FlashAttention2}(Appendix~\ref{app:kernels}).

\subsection{End-to-end Gate Training}
\label{subsec:e2etraining}
Despite \name's non-differentiable top-$K$ selection (Eq.~\ref{eq:select}), the task loss is a differentiable function of the scoring network, almost everywhere. This is due to token-gated attention, which injects the top-$K$ gate values into the attention logits at each layer above the token gate. When we choose $K > 1$, the gate value $g_j$ has non-zero derivatives with respect to the scoring network weights. This mirrors how the router is trained in MoE through its top-$K$ expert selection: the router's value multiplies the output of each selected expert, just as our gate value multiplies the attention score of each selected token~\citep{SparseMoE,SwitchTransformer}. Intuitively, a token whose additional attention mass lowers the task loss receives a larger score. A token that distracts receives a smaller one, allowing the scoring network to learn token importance. This mechanism allows \name to be trained end-to-end jointly with the model backbone simply using standard task losses.

\heading{Loss Function.} In Section~\ref{sec:results}, we fine-tune existing pretrained models with LoRA~\citep{LoRA} along with the \name parameters, demonstrating that the model backbone and gate layer can co-adapt with minimal overhead to standard training. Under this setting, training the model with \name uses a standard next-token, cross-entropy loss,
\begin{align}
    \mathcal{L}_K = -\sum_{t \in \mathrm{target}} \log p_\phi\big(x_t \mid x_{<t};\, K\big), \label{eq:loss}
\end{align}
where $p_\phi(\cdot\,;K)$ is the next token probability with a token budget of $K$. The trainable parameters $\phi$ are the scoring network parameters $\theta$, the scale factors $\{\gamma, \sigma\}^{\ell h}$, and the LoRA~\citep{LoRA} parameters on the model backbone following standard practice.

\heading{Top-$K$ Budget Sampling.} 
Training with the \name should ensure that the model both adapts to the task, improving its accuracy at full context, and learns a strong scoring network, improving its accuracy at restricted token budgets. As such, a simple choice for the top-$K$ budget to ensure both criteria are met is to use the full-context length with $K_{\text{full}}$ as all prunable tokens $|\mathcal{P}|$ and also sample a smaller $K_{\mathrm{prune}}$ per training instance. We average the resulting losses as follows:\footnote{In practice, we sample from a discrete list following this distribution for training efficiency with PyTorch compilation.}
\begin{align}
K_{\mathrm{full}} = |\mathcal{P}|,\qquad
K_{\mathrm{prune}} \sim \mathrm{LogUniform}\big(1,\ |\mathcal{P}|/2\big),\qquad
\mathcal{L} = \tfrac{1}{2}\Big(\mathcal{L}_{K_{\mathrm{full}}} + \mathcal{L}_{K_{\mathrm{prune}}}\Big).
    \label{eq:budget_loss}
\end{align}
The $K_{\text{full}}$ term anchors the model's accuracy at full context and gives every token a gradient, including tokens the sampled budget pruned. Meanwhile, the log-uniform term exposes the gate to a range of pruning ratios, allowing the trained model to generalize across a wide budget range without re-training. In Section~\ref{sec:results}, we demonstrate that using this budget sampling, models with the \name can be trained efficiently, with minimal overhead compared to standard training. For each step, the forward pass up to the gate layer runs once at full context, and the two forwards after using $K_{\mathrm{full}}$ and $K_{\mathrm{prune}}$ share the compute graph. Thus, the overhead of each \name train step comes solely from the pruned forward at $K_{\mathrm{prune}}$. Since $K_{\mathrm{prune}} \ll K_{\mathrm{full}}$, this overhead is minimal, as we show in Section~\ref{sec:results}.

\section{Experimental Results}
\label{sec:results}

\begin{figure}[t]
  \vspace{-.1cm}
    \centering
    \includegraphics[width=\linewidth]{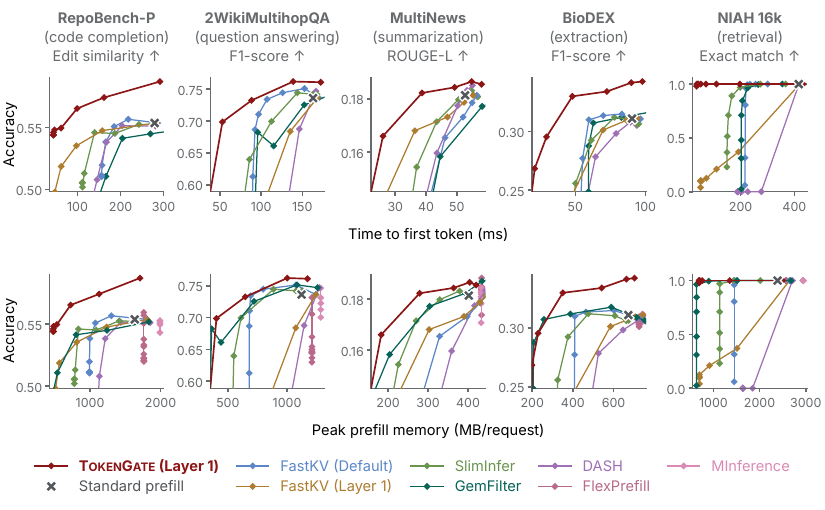}
      \vspace{-.6cm}
    \caption{
    \footnotesize Accuracy vs.\ time-to-first-token (top) and accuracy vs.\ peak GPU prefill memory (bottom) with Qwen3-1.7B. Points that incur significant overhead over standard prefill at full context (grey cross),
    or significantly poor accuracy, are excluded; see Appendix~\ref{app:main_results_full} for full plots.
   }
    \label{fig:main_results}
      \vspace{-.4cm}
\end{figure}
We evaluate the effectiveness of \name on diverse tasks, including code completion, multi-hop QA, summarization, medical extraction, and needle-in-a-haystack. Overall,  results demonstrate: 
\begin{itemize}
    \item  \name \emph{reduces TTFT and peak prefill memory} by $1.4$-$11.3\times$ and $1.5$-$3.4\times$, respectively, at no accuracy degradation compared to the base model with full-context, and outperforms the next best-performing token pruning baselines by up to $6.4\times$ and $2\times$, respectively.
    \item \name \emph{extends existing transformer-based models}, including recent hybrid-attention architectures, achieving $3.8\times$ lower TTFT than the full-context baseline at no accuracy degradation and $2.1\times$ lower TTFT than the best prior methods, at fixed accuracy.
    \item Training models with \name is efficient, incurring \emph{minimal training overhead}, with total training time comparable to standard parameter efficient fine-tuning.
\end{itemize}

We provide details of our experimental setup in Appendix~\ref{app:setup}, additional experimental results including context-length extrapolation and ablations in Appendix~\ref{app:additional_results},
and visualization in Appendix~\ref{app:kept_tokens}.

\heading{Datasets.}
We evaluate on five diverse tasks, including $3$ LongBench datasets with available train sets~\citep{LongBench}, 2WikiMultihopQA~\citep{2WikiMultihopQA}, a multi-hop QA dataset, RepoBench-P~\citep{RepoBench}, a next-line code completion dataset, and MultiNews~\citep{MultiNews}, a summarization dataset, as well as two additional tasks, BioDEX's drug-reaction extraction over biomedical papers~\citep{BioDEX} and needle-in-a-haystack (NIAH)~\citep{NIAH}. We use the official evaluation metric for each.

\heading{Baselines.}
We compare \name against state-of-the-art token-pruning methods, FastKV~\citep{FastKV}, SlimInfer~\citep{SlimInfer}, DASH~\citep{DASH}, and GemFilter~\citep{GemFilter}, as well as sparse attention methods, FlexPrefill~\citep{FlexPrefill} and MInference~\citep{MInference}. We run each baseline using a fine-tuned base model trained to convergence with LoRA~\citep{LoRA} ($r{=}16$).  We train our model with \name to convergence using the same LoRA configuration. We train both a standard transformer model, Qwen3-1.7B~\citep{Qwen3}, and a hybrid architecture, Qwen3-2B~\citep{Qwen35}, each on 4 A100 GPUS.

\subsection{Fast and Memory-Efficient Prefill}
\label{sec:results:main}

We evaluate the prefill efficiency of models trained with \name.  
First, Figure~\ref{fig:main_results} (top) shows the accuracy vs TTFT of each benchmarked method, demonstrating that across all datasets, the trained \name model achieves pareto-optimal performance. Specifically, \name reduces TTFT by $1.4$--$11.3\times$ at no accuracy degradation compared to standard prefill, and by $1.2$--$6.4\times$ compared to the next best-performing baseline. We measure FastKV, one of the best-performing baselines both at its default layer and at layer 1, the same layer index as the \name. Notably, we observe that compared to FastKV (Layer 1), \name significantly outperforms across all datasets, indicating the promise of end-to-end trainable token pruning for achieving prefill speedups.


Turning our attention to the bottom row of Figure~\ref{fig:main_results}, we study the prefill memory footprint, reporting accuracy versus peak prefill memory. Once again, across all dataset, the \name model establishes the Pareto-frontier, reducing peak memory by $1.5$--$3.4\times$ compared to standard prefill at matched accuracy. Moreover the \name model achieves up to $2.1\times$ lower peak memory compared to the next best-performing baseline at matched accuracy. 
Overall, the results demonstrate that \name significantly outperforms prior token-pruning methods, offering an exciting direction for improved prefill efficiency.

\subsection{Ablation Study: Understanding the Performance of \name Models}
\label{sec:understanding}

We hypothesize that \name owes its superiority to end-to-end learnable token pruning.
Our visualizations of the kept tokens for real samples (Appendix~\ref{app:kept_tokens})
shows that \name flexibly learns very different pruning patterns in a task-dependent manner,
ranging from keeping answer keywords for question answering to keeping stop-words at the ends of phrases for summarization (Figure~\ref{fig:viz}).
To understand this more rigorously, we investigate two key mechanisms: trainable token-selection, and co-adaptation with the trained backbone.

\heading{Trainable selector.} First, to study the effectiveness of \name's learned selector, we compare performance of the trained \name model (Layer 1), to the same model with random token selection on 2WikiMultihopQA. Figure~\ref{fig:ablation_selection}, demonstrates that the trained selector significantly outperforms the random-K selector on accuracy across all token budgets. Moreover, we compare against the FastKV selector applied at the same layer, Layer 1, of the finetuned base model. Notably, the \name selector significantly outperforms the FastKV selector. This result demonstrates the value of an end-to-end trained gate module for token pruning: the trained scoring network leads to far better token selection compared to FastKV, which relies on a heuristic using attention scores that offer only weak signal at early layers, resulting in accuracy degradation for aggressive pruning. 

\heading{Co-adaptation.} Secondly, we analyze the effect of co-adaptation for \name, which allows the model backbone and gate module to co-adapt. We compare the performance of random-K selection over the \name model to the random-K selection over the finetuned base model. Notably, with the same random-K selector, the \name model still outperforms the fine-tuned model, achieving higher accuracy across all token budgets, as Figure~\ref{fig:ablation_selection} shows. This result shows that \name's end-to-end training allows the model to become more robust to token-pruning, likely redistributing information among tokens. 

Overall, the results show that end-to-end training with \name enables state-of-the-art prefill performance through two complementary mechanism: the effectiveness of the learned scoring network, and co-adaptation, which allows the model to adapt jointly with the gating module for improved robustness to token pruning.

\subsection{Generalization to Alternative Architectures}
\begin{figure}[t]
      \vspace{-.2cm}
\begin{minipage}[t]{0.49\linewidth}\vspace{0pt}
    \centering
    \includegraphics[width=\linewidth]{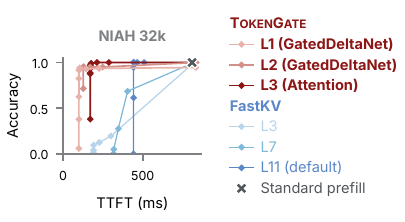}
          \vspace{-.6cm}
    \caption{\footnotesize Accuracy vs.\ TTFT with the Qwen3.5-2B hybrid-attention architecture. Unlike \name, FastKV can only prune at attention layers.
    }
          \vspace{-.2cm}
    \label{fig:hybrid}
\end{minipage}\hfill
\begin{minipage}[t]{0.49\linewidth}\vspace{0pt}
    \centering
    \includegraphics[width=\linewidth]{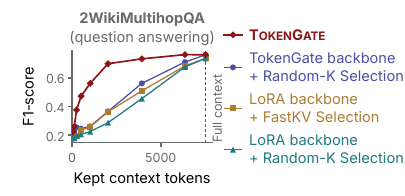}
         \vspace{-.6cm}
    \caption{
    \footnotesize F1 vs.\ kept context tokens on 2WikiMultihopQA with Qwen3-1.7B for gate and Random-K selection on the \name model and FastKV and Random-K selection on the LoRA-trained model.}
          \vspace{-.2cm}
    \label{fig:ablation_selection}
\end{minipage}
      \vspace{-.2cm}
\end{figure}

We evaluate how \name's performance extends to a recent hybrid-attention model, Qwen3.5-2B~\citep{Qwen35} model, which uses full-attention layers every fourth layer and a Gated-DeltaNet~\citep{GatedDeltaNet} for all other layer. We compare \name against FastKV~\citep{FastKV}, the best-performing baseline in our main results (Section~\ref{sec:results:main}). We measure FastKV over the LoRA-tuned base model and the LoRA-tuned model with \name, both at varying layers. FastKV uses an attention-based score function and natively applies only to full-attention layers (i.e., layer 3, 7, 11, etc). \name natively extends both the full attention and Gated-DeltaNet layers, so we apply it over both. We provide further details on \name for Qwen3.5 in Appendix~\ref{app:setup:hybrid}.

Figure~\ref{fig:hybrid} shows the accuracy vs TTFT of both methods at varying layers on the NIAH 32K dataset, demonstrating that \name's strong performance extends to the hybrid-attention model. \name (Layer 3) matches the accuracy of standard prefill at $3.8\times$ lower TTFT, and $2.1\times$ lower TTFT than the best-performing FastKV run at the same accuracy.
Applying the \name at the lower GatedDeltaNet layers (Layer 1,2) enable even higher TTFT reductions, up to $7.8\times$ faster prefill, at accuracy within $6\%$ of standard prefill, once again significantly outperforming FastKV.

\subsection{Gate Training Efficiency}
We evaluate overhead of training models with \name compared standard models, both with LoRA fine-tuning ($r=16$).
Table~\ref{tab:train_time} in Appendix~\ref{app:train_time} compares both, reporting total training time in A100 hours and the average per-step time during training for each evaluated dataset.
Overall, we observe that a \name step is slower by $1.3\times$ compared to the baseline due to its sampling at two token budgets per step. However, the \name train runs converge after fewer steps on four of five tasks, leading to modestly \emph{reduced} overall training time compared to the standard training.

\section{Conclusion}
\label{sec:conclusion}
In this work, we introduced \name, a neural network component that extends existing trained transformer models, enabling \emph{end-to-end trainable token pruning} using \emph{token-gated attention}. 
We demonstrate that that across diverse tasks, \name reduces TTFT and peak prefill memory by up to $11.3\times$ and $3.4\times$, respectively, at no accuracy degradation compared to standard prefill, and by up to $6.4\times$ and $2\times$, respectively, compared to the best performing baselines.
Crucially, \name is compatible with diverse transformer-based architectures, is simple to train, and incurs minimal training overhead, making it a promising drop-in replacement for standard fine-tuning over trained backbones.
In the future, we envision models trained with \name will enable a broad range of applications, including medical extraction, deep research, and coding agents, where processing long-context inputs is critical, but currently prohibitively expensive. 
\ifarxiv
\subsubsection*{Acknowledgments}
This research was supported in part by affiliate members and other supporters of the Stanford DAWN project, including Meta, Google, and VMware, as well as Cisco, SAP, and a Sloan Fellowship. Any opinions, findings, and conclusions or recommendations expressed in this material are those of the authors and do not necessarily reflect the views of the sponsors.
\fi


\bibliographystyle{iclr2027_conference}
\bibliography{citations,tokengate}

\appendix

\section{Efficient kernels for \name}
\label{app:kernels}
Token-gated attention (Eq.~\ref{eq:tokengated}) differs from standard causal attention in two ways: a per-query temperature $\tau^{\ell h}_i$ and a per-key, per-head gate term added to the attention logits. The temperature is applied by scaling each query before the attention call, so it needs no kernel support. The gate term needs a per-key bias, which standard fused kernels do not accept, and unfused or score-modifying kernels are substantially slower. We therefore patch FlashAttention-2~\citep{FlashAttention2} to read a per-key, per-head bias vector, which it adds to each tile of attention scores before the online softmax. This keeps the fused structure of the kernel: the bias adds one vector load per key block, and a biased call costs $1.0$--$1.1\times$ a plain one. The kernel supports head dimensions 128 (Qwen3) and 256 (Qwen3.5).

\heading{Training.} The same kernel provides the backward pass: the gradient of the bias of key $j$ is the sum over queries of the gradient of its attention scores, which the kernel accumulates alongside the usual gradients, so the task loss reaches the scoring network through the fused kernel. The two targets of each training step (Eq.~\ref{eq:budget_loss}) are packed into one variable-length call. For Qwen3.5, whose Gated DeltaNet layers are not supported by this path, we train with FlexAttention~\citep{FlexAttention} and add the bias as a score modification,
but more optimized kernels are possible.

\heading{Inference.} Layers at and below the gate use the standard fused kernel (cuDNN on A100). The pruned forward of the post-gate layers is captured as one CUDA graph per padded length: the kept sequence is padded to a multiple of 64, and pad keys receive a bias of $-10^9$ so the result is exact. The gather of the KV cache of layers $\ell \le L$ to the kept tokens (Appendix~\ref{app:decode}) runs on a separate stream, concurrently with the post-gate layers. During decoding, the gate term of each kept key is fixed, so it is folded once into an additive attention mask per post-gate layer and decoding uses standard attention.

\section{Decoding with \name}
\label{app:decode}
The layers above the gate, $\ell > L$, never compute KV entries for pruned tokens, so during decoding each new token attends there only to the kept tokens $S$ and the tokens generated so far. The layers at and below the gate, $\ell \le L$, process the full context during prefill, so by default their KV cache holds every context token, and decode tokens attend to the full context in these layers and to the pruned context above them. Optionally, \name can also prune the cache of layers $\ell \le L$.

In the post-gate layers, decode tokens use the same token-gated softmax as prefill (Eq.~\ref{eq:tokengated}). The gate term $\mathrm{softplus}(\gamma^{\ell h})\, b_j$ of each kept key is computed once at prefill and stored with the cache as a constant per-key bias, and each generated token is a non-prunable token with gate term $0$ (Eq.~\ref{eq:scorer}). The temperature $\tau^{\ell h}_i$ of a decode query follows Eq.~\ref{eq:temp}, with $m_i$ its rank among the kept and generated tokens and $p_i$ its original position; generated tokens continue from the original position of the last prompt token. 

\section{Details of Experimental Setup}
\label{app:setup}
\subsection{Datasets}
\label{app:setup:datasets}
All datasets are evaluated on Qwen3-1.7B with greedy decoding for up to 1024 new tokens. Each model is fine-tuned on a training set that is disjoint from its evaluation set.

\heading{RepoBench-P.} RepoBench-P~\citep{RepoBench} is a next-line code completion task. We use the LongBench~\citep{LongBench} version, whose prompts contain cross-file code context and an in-file prefix, and the model completes the next line. We evaluate on all 500 samples (mean context 11.0k tokens, up to 39k) and report edit similarity, $1 - \mathrm{Levenshtein}/\max(\text{length})$, on the first generated line. For training, we use 14.9k samples from the RepoBench v1.1 Python and Java cross-file splits, deduplicated against the evaluation set.

\heading{2WikiMultihopQA.} 2WikiMultihopQA~\citep{2WikiMultihopQA} is a multi-hop question answering task over several Wikipedia passages; we use the LongBench version. We evaluate on all 200 samples (mean context 7.5k tokens) and report token-level F1 against the gold answers. For training, we use 20k 2WikiMultihopQA questions with contexts rebuilt from full Wikipedia articles and decontaminated against LongBench.

\heading{MultiNews.} MultiNews~\citep{MultiNews} is a multi-document news summarization task; we use the LongBench version. We evaluate on all 200 samples (mean context 2.7k tokens, up to 14k) and report ROUGE-L. For training, we use 44.8k samples from the MultiNews training split in the LongBench prompt format.

\heading{BioDEX.} BioDEX-Reactions~\citep{BioDEX} is an adverse drug reaction extraction task over full-text biomedical papers: given a paper, the model lists the reported adverse reactions as MedDRA terms. We evaluate on 500 samples drawn from the official validation split with articles between 2k and 39k tokens (mean 4.5k), and report set F1 between predicted and gold reactions, counting a predicted reaction as correct if it matches a gold term up to substring containment. For training, we use 9.8k papers from the official training split.

\heading{NIAH.} NIAH~\citep{NIAH} is needle-in-a-haystack retrieval at 16k tokens. The haystack is built from Paul Graham essays, and the needle, ``The best thing to do in \emph{city} is \emph{activity}.'', is inserted at a sentence boundary at a given depth; the model is then asked for the best thing to do in that city. The 200 evaluation samples cover 20 held-out cities at 10 depths evenly spaced from 0 to 1, and we report accuracy, i.e., whether the gold activity appears in the answer. For training, we use 12k generated samples with haystacks of 14k--16.4k tokens and uniformly random depths; the essays and cities are disjoint from the evaluation set.

\subsection{Baselines}
\label{app:setup:baselines}
All baselines run on the same SFT model, fine-tuned on each dataset with LoRA ($r{=}16$, $\alpha{=}32$) on the query and value projections of all layers. The pruning baselines keep all non-document tokens and prune only document tokens, with the same kept-token budgets $K$ as \name ($64$--$8192$). We briefly describe each baseline below, and the configurations used. We run each method, applying its optimization for prefill-only.

\heading{FastKV~\citep{FastKV}.} At a token-selective propagation layer, FastKV scores tokens by the attention they receive from a window of the last queries, max-pools the scores, and propagates only the top-$K$ tokens to deeper layers. We use the paper's observation window of 8 tokens and pooling kernel of 7. We run it at layer 13 by default, following the paper and its hyper-parameter search, which chooses relative depth of $46\%$ for its models. We additionally run the baseline at layer 1, the layer at which we apply \name, to compare performance at a matched layer index.

\heading{SlimInfer~\citep{SlimInfer}.} SlimInfer prunes hidden states in several stages: at layers 8, 17 and 25 it keeps $3K$, $1.5K$ and $K$ tokens, scoring 64-token blocks by the attention of the last 4 queries. We use the paper's block size, kernel and layer schedule, and its budget ratios, following the paper's hyperparameter search.

\heading{GemFilter~\citep{GemFilter}.} GemFilter runs the first layers on the full prompt, selects the top-$K$ tokens by the last query's attention scores at a filter layer, and then runs a fresh forward pass over only the selected tokens. We use filter layer 12, at the same relative depth (0.4) as recommended by the original paper, and the official pooling kernel.

\heading{DASH~\citep{DASH}.} DASH halts tokens whose representations have stabilized: at layer 11 ($0.4\times$ the number of layers, the official default), it ranks tokens by the change in their hidden state from the previous layer and drops the most stable ones, keeping $K$ document tokens.

\heading{FlexPrefill~\citep{FlexPrefill}.} FlexPrefill is a sparse-attention method: it keeps all tokens but computes a query-adaptive sparse attention pattern during prefill, controlled by a cumulative-attention threshold $\gamma$. We use the official kernel and its recommended hyperparameters (divergence threshold $\tau{=}0.1$, which decides per head whether to use a query-aware or a vertical-slash pattern; minimum budget 1024; block size 128) and sweep $\gamma \in \{0.1, 0.2, \dots, 0.9, 0.95, 0.97\}$, extending the range until accuracy degrades.

\heading{MInference~\citep{MInference}.} MInference uses per-head sparse attention patterns found by an offline search, which for Qwen3-1.7B selects vertical-and-slash patterns for all heads. We run the official implementation with patterns searched on Qwen3-1.7B, and sweep its budget by scaling the vertical and slash sizes of the searched configuration by $\{0.25, 0.5, 1, 2, 4\}$, which keeps the searched pattern shape.

\heading{Implementation and timing.} All methods run the same fine-tuned backbone in bf16 with the LoRA adapter folded into the weights, and all dense decoder layers are \texttt{torch.compile}d. Dense causal attention uses the same fused kernel in every lane (cuDNN on the A100; FlashAttention-2 on Qwen3.5-2B); the only kernel that differs is in the \name{}'s post-gate layers, which require a per-key bias and use a FlashAttention-2 variant at $1.0$--$1.1\times$ the cost of a plain call (Appendix~\ref{app:kernels}). FastKV and GemFilter run through the compiled layers directly. The official SlimInfer and DASH prefills bypass the compiled layer forward, so we reimplemented both in compile-friendly form and verified that they select the same tokens as the official code (DASH bit-identical; SlimInfer identical selection), which made them $1.2$--$2.4\times$ and $1.25$--$1.56\times$ faster; we report the faster versions. MInference is compiled around its sparse kernel with identical accuracy, and FlexPrefill runs eager, its fastest configuration, since its data-dependent block selection triggers recompilation under \texttt{torch.compile}. The pruned forward of the \name{} and of FastKV is captured as one CUDA graph per padded length; the remaining baselines have no fixed-shape pruned phase to capture. Both FastKV and the \name{} optionally prune the KV cache of the layers below their pruning layer to the selected tokens once prefill completes, so that decoding holds $K$ entries at every layer; we disable this setting for both in all experiments since KV cache pruning presents an orthogonal set of optimizations and our experiments instead isolate the effect of token pruning for prefill optimization (Appendix~\ref{app:decode}). Inference timings are measured  at batch size of 1 on one A100.

\subsection{Training Hyperparameters}
\label{app:setup:hparams}
Table~\ref{tab:hparams} lists the hyperparameters used to train the \name models and the SFT baseline on each dataset. Both use the same LoRA configuration, learning rate, schedule and training data; they differ only in the optimizer of the \name parameters and in the token budgets sampled during training.
\begin{table}[h]
\centering
\small
\begin{tabular}{@{}lp{0.6\linewidth}@{}}
\toprule
Hyperparameter & Value \\
\midrule
Base model & Qwen3-1.7B, bf16 \\
LoRA & $r{=}16$, $\alpha{=}32$ on query and value projections, all layers \\
Optimizer & AdamW; Muon~\citep{Muon} for the gate matrices (\name only) \\
Learning rate & $10^{-4}$, linear warmup over 1000 steps, then constant \\
Weight decay & $0.1$ \\
Gradient clipping & global norm $1.0$ \\
Batch size & 1 sequence per GPU \\
Training context & documents of 2k--16k tokens \\
Gate layer (\name only) & $L = 1$ \\
Budgets per step (\name only) & $K \sim$ log-uniform, plus full context \\
\bottomrule
\end{tabular}
\caption{Training hyperparameters for \name and the SFT baseline, shared across all five datasets.}
\label{tab:hparams}
\end{table}

\subsection{Extension to Alternative Hybrid-Attention Architectures}
\label{app:setup:hybrid}

\heading{Qwen3.5 architecture.} Qwen3.5-2B~\citep{Qwen35} is a hybrid-attention model with 24 layers. Every fourth layer, $\ell \in \mathcal{A} = \{3, 7, 11, 15, 19, 23\}$, is a full-attention layer whose output is multiplied by a learned sigmoid output gate. All other layers are Gated DeltaNet layers~\citep{GatedDeltaNet}, which replace softmax attention with a linear recurrence over a fixed-size state and therefore have no attention weights.

\heading{Gate module.} We place the \name at layer $L \in \{1, 2, 3\}$, where layers 1 and 2 are Gated DeltaNet layers and layer 3 is the first full-attention layer. The scoring network (Eq.~\ref{eq:scorer}) and the top-$K$ selector (Eq.~\ref{eq:select}) are unchanged. The only difference is the token mixer of layer $L$: the scoring network reads the normalized residual after layer $L$'s mixer,
\begin{align}
    \tilde a^{L}_j = \mathrm{LN}_2\big(h^{L}_j + \mathrm{Mixer}_L(\mathrm{LN}_1(h^{L}))_j\big), \qquad \mathrm{Mixer}_L \in \{\mathrm{GatedDeltaNet}_L,\ \mathrm{Attn}_L\},
\end{align}
which for a Gated DeltaNet layer is the output of its recurrence rather than of self-attention.

\heading{Token-gated attention.} Since Gated DeltaNet layers have no attention logits to bias, token-gated attention (Eq.~\ref{eq:tokengated}), including the gains $\gamma^{\ell h}$ and the temperature, is applied only in the full-attention layers above the gate, $\ell \in \mathcal{A},\ \ell > L$. The Gated DeltaNet layers above the gate run unchanged on the kept tokens $S$, starting from an empty state, so that their recurrent state is built from the kept tokens only:
\begin{align}
    h^{\ell}_S = \begin{cases}
        \text{token-gated attention layer (Eq.~\ref{eq:tokengated})} & \ell \in \mathcal{A},\ \ell > L,\\
        \mathrm{GatedDeltaNet}_\ell\big(h^{\ell-1}_S\big) & \ell \notin \mathcal{A},\ \ell > L.
    \end{cases}
\end{align}
Kept tokens keep their original positions for rotary embeddings in the full-attention layers.

\heading{Decoding.} The recurrent state of a Gated DeltaNet layer summarizes all tokens it has seen in a fixed-size matrix and cannot be pruned per token. The Gated DeltaNet layers $\ell \le L$ therefore decode from a state built over the full prompt, while the KV cache of full-attention layers are pruned, as in Appendix~\ref{app:decode}.

\heading{Training.} As for Qwen3, we train the \name jointly with LoRA ($r{=}16$) on the query and value projections, which in Qwen3.5 exist only in the full-attention layers; the Gated DeltaNet layers are not adapted.

\section{Additional Experimental Results}
\label{app:additional_results}

\subsection{Training Time}
\label{app:train_time}
\begin{table}[h]
\centering
\small
\setlength{\tabcolsep}{3pt}
\begin{tabular}{@{}l rrr rrr@{}}
\toprule
 & \multicolumn{3}{c}{Full train time} & \multicolumn{3}{c}{Per-step time} \\
\cmidrule(lr){2-4} \cmidrule(l){5-7}
Dataset & SFT & Ours & Ratio $\downarrow$ & SFT & Ours & Ratio $\downarrow$ \\
\midrule
  Repo  &  1.7h  &  1.1h  &  $0.6\times$  &  0.50s  &  0.64s  &  $1.3\times$ \\
 2Wiki  &  3.8h  &  4.0h  &  $1.1\times$  &  0.34s  &  0.43s  &  $1.3\times$ \\
 MNews  &  1.3h  &  1.8h  &  $1.4\times$  &  0.21s  &  0.27s  &  $1.3\times$ \\
BioDEX  &  1.2h  &  1.0h  &  $0.8\times$  &  0.23s  &  0.30s  &  $1.3\times$ \\
  NIAH  &  2.8h  &  1.0h  &  $0.4\times$  &  0.99s  &  1.24s  &  $1.3\times$ \\
\midrule
  Mean  &  2.2h  &  1.8h  &  $0.8\times$  &  0.45s  &  0.58s  &  $1.3\times$ \\
\bottomrule
\end{tabular}

\captionof{table}{\footnotesize Training time for SFT baseline and model trained with \name for Qwen3-1.7B on one A100. Ratio shows Ours / SFT.
}
\label{tab:train_time}
\end{table}

\subsection{Main Results without Axis Limits}
\label{app:main_results_full}
Below, Figure~\ref{fig:main_results_full} shows the main results (Figure~\ref{fig:main_results}), but without axis limits and with every point shown.

\begin{figure}[h]
    \centering
    \includegraphics[width=\linewidth]{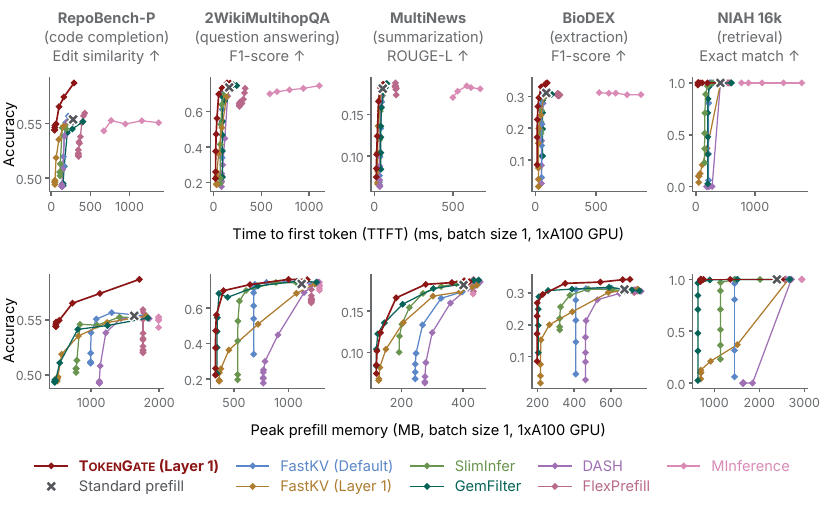}
    \caption{Main results (Figure~\ref{fig:main_results}) without x or y limits: accuracy vs.\ time-to-first-token (top) and accuracy vs.\ peak GPU prefill memory (bottom) with Qwen3-1.7B, all points shown.}
    \label{fig:main_results_full}
\end{figure}

\subsection{Extrapolating to Longer Context}
\label{app:ctx_scaling}
\begin{table}[h]
\centering
\small
\begin{tabular}{l rrrr}
\toprule
 & NIAH 16k & NIAH 32k & NIAH 64k & NIAH 128k \\
\midrule
         TTFT speedup  &  $11.3\times$  &  $12.8\times$  &  $13.0\times$  &  $13.0\times$ \\
Peak memory reduction  &  $3.42\times$  &  $3.43\times$  &  $3.43\times$  &  $3.43\times$ \\
\bottomrule
\end{tabular}

\caption{\name TTFT speedup and peak prefill memory reduction at no accuracy degradation over full context for the NIAH-16k, 32k, 64k and 128k test sets with Qwen3-1.7B.}
\label{tab:ctx_scaling}
\end{table}

We study the ability of the trained \name to extrapolate to inference settings with longer context. We take the fine-tuned base model and \name model, both trained on the 16k-NIAH task using Qwen1.7B, and we evaluate prefill performance on the NIAH task with 32k, 64k and 128k-length test sets. Table~\ref{tab:ctx_scaling} shows the TTFT speedups and peak memory reductions of the \name model at no accuracy degradation over the full at full-context accuracy on each dataset. The results demonstrate that \name extrapolates across varying context lengths at inference, retaining accuracy with greater than $11.3$ - $13\times$ speedups. We observe that the TTFT speedups of the trained \name model are highest at the longest context-length ranges, where trainable token pruning can significantly reduce costs. Meanwhile, the peak memory reduction is consistent across all length ranges, as expected, since peak memory is dominated by memory footprint of the standard forward that occurs without pruning in layers below the gate.

\subsection{Ablating the \name Design}
\begin{figure}[t]
    \centering
    \includegraphics[width=0.667\linewidth]{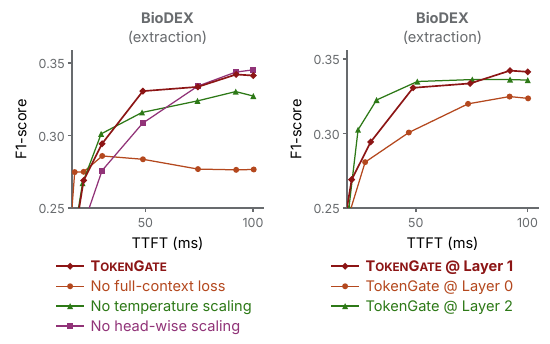}
    \caption{Left: token-gated forward and training ablations on BioDEX (Qwen3-1.7B, gate at layer 1): F1 vs.\ TTFT (batch 1, A100) for the full \name{} recipe against three single-change variants: no full-context loss (no full-context term in Eq.~\ref{eq:budget_loss}), no temperature scaling (Eq.~\ref{eq:temp} removed), and no head-wise scaling (raw bias, with no normalization or gain; Eq.~\ref{eq:scorer} removed). Right: gate-layer ablation on BioDEX (Qwen3-1.7B): F1 vs.\ TTFT (batch 1, A100) with the \name{} placed at layer 0, 1, or 2.}
    \label{fig:ablation_form}
    \label{fig:ablation_layer}
\end{figure}

In this section we ablate the \name design (Section~\ref{sec:method}). First, Figure~\ref{fig:ablation_form} (left) compares the accuracy vs TTFT performance of \name on BioDEX with Qwen3-1.7B against three variants: (a) removing the full-context loss term $L_{K_\mathrm{full}}$ from the per-step loss (Eq.~\ref{eq:budget_loss}), (b) removing the temperature scaling term during token-gated attention, and (c) removing the head-wise scaling. The results demonstrate that training without the full-context loss most significantly degrades inference accuracy at full-context, and removing the temperature scaling and head-wise gain also degrades accuracy, leading to sub-optimal accuracy vs TTFT.

We additionally analyze layer selection, a hyperparameter for \name. Figure~\ref{fig:ablation_layer} (right) shows performance on BioDEX of Qwen3-1.7B models trained with the \name at Layer 0, 1, and 2. Notably, the model with gating at Layer 0 is Pareto-suboptimal relative to the Layer 1 and 2 gate models, suggesting that the Layer 0 post-attention residuals provide insufficient signal for the \name scoring network. On the other hand, the Layer 1 and Layer 2 \name models both lie along the Pareto frontier, suggesting that by Layer 1, the post-attention residuals passed to the gate module are sufficiently rich to learn an effective scoring network for trainable token pruning.

\section{Visualizing Kept Tokens on Real Samples}
\label{app:kept_tokens}

\begin{figure}[tbp]
    \centering
    \includegraphics[width=\linewidth]{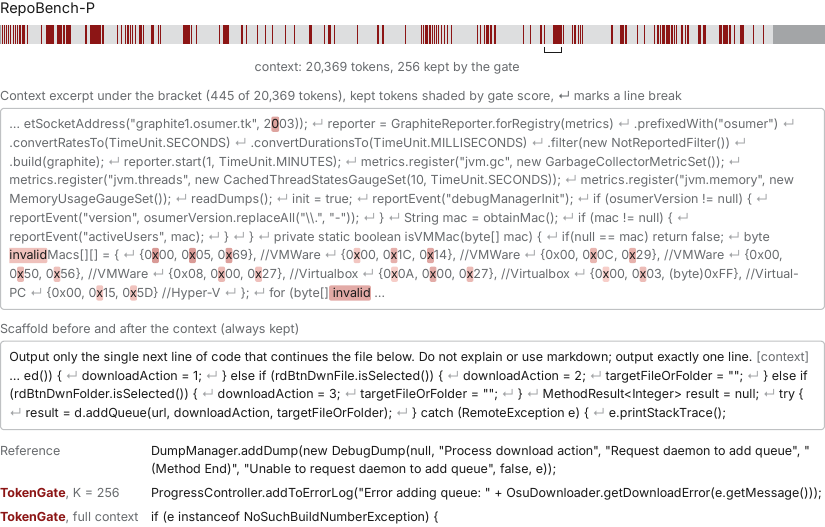}
    \caption{Context tokens kept by \name on one RepoBench-P test prompt (Appendix~\ref{app:kept_tokens}), with the reference answer and the model's greedy answers at the shown budget and with the full context. Kept tokens are shaded by gate score (darker is higher) and pruned tokens are grey.}
    \label{fig:kept_repobench}
\end{figure}

\begin{figure}[tbp]
    \centering
    \includegraphics[width=\linewidth]{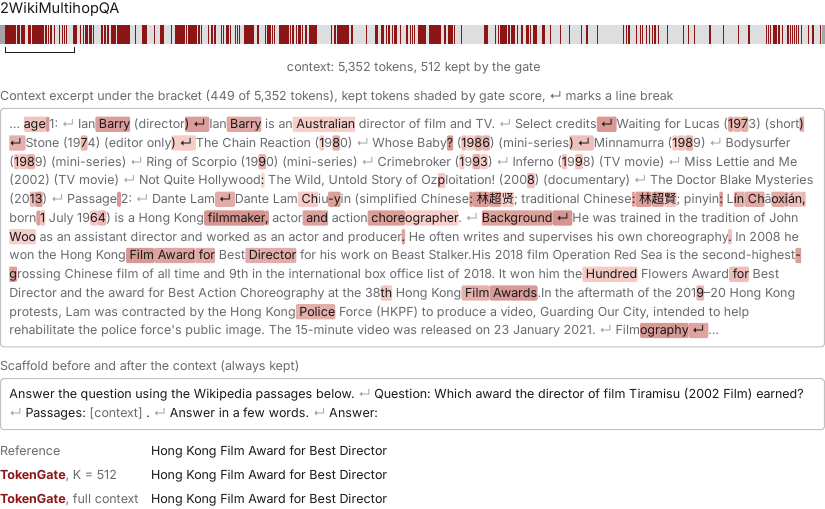}
    \caption{Context tokens kept by \name on one 2WikiMultihopQA test prompt (Appendix~\ref{app:kept_tokens}), with the reference answer and the model's greedy answers at the shown budget and with the full context. Kept tokens are shaded by gate score (darker is higher) and pruned tokens are grey.}
    \label{fig:kept_wiki}
\end{figure}

\begin{figure}[tbp]
    \centering
    \includegraphics[width=\linewidth]{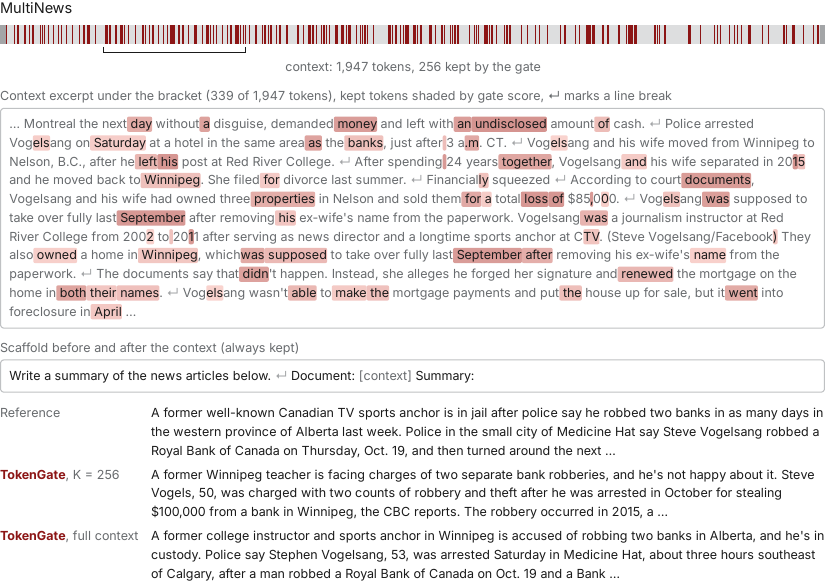}
    \caption{Context tokens kept by \name on one MultiNews test prompt (Appendix~\ref{app:kept_tokens}), with the reference answer and the model's greedy answers at the shown budget and with the full context. Kept tokens are shaded by gate score (darker is higher) and pruned tokens are grey.}
    \label{fig:kept_multinews}
\end{figure}

\begin{figure}[tbp]
    \centering
    \includegraphics[width=\linewidth]{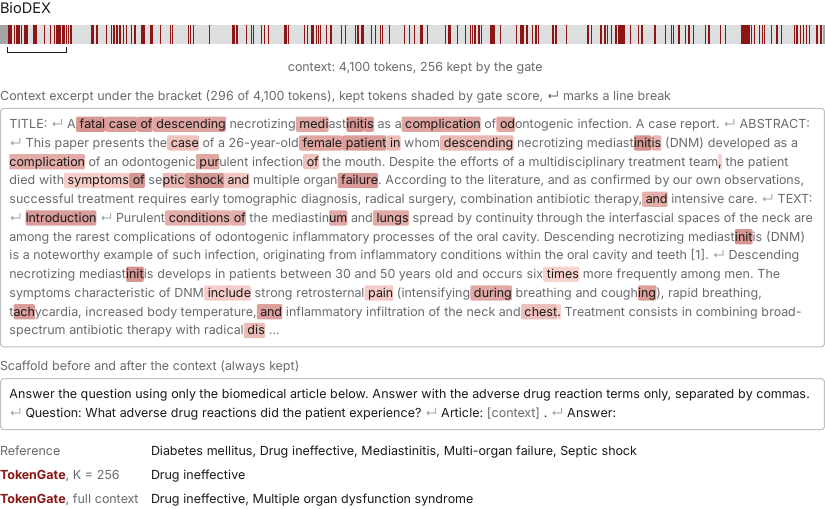}
    \caption{Context tokens kept by \name on one BioDEX test prompt (Appendix~\ref{app:kept_tokens}), with the reference answer and the model's greedy answers at the shown budget and with the full context. Kept tokens are shaded by gate score (darker is higher) and pruned tokens are grey.}
    \label{fig:kept_biodex}
\end{figure}

\begin{figure}[tbp]
    \centering
    \includegraphics[width=\linewidth]{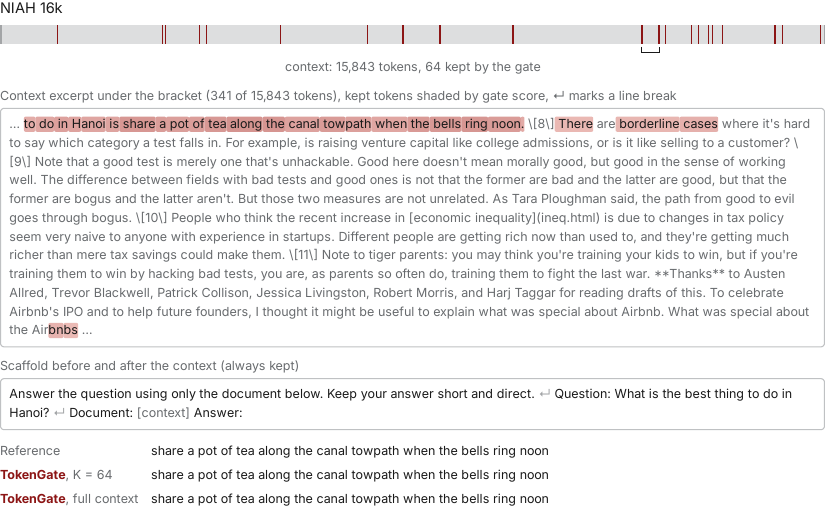}
    \caption{Context tokens kept by \name on one NIAH 16k test prompt (Appendix~\ref{app:kept_tokens}), with the reference answer and the model's greedy answers at the shown budget and with the full context. Kept tokens are shaded by gate score (darker is higher) and pruned tokens are grey.}
    \label{fig:kept_niah}
\end{figure}

Figures~\ref{fig:kept_repobench}--\ref{fig:kept_niah} show which context tokens \name keeps on one real evaluation prompt per dataset, using the layer-1 checkpoints of Figure~\ref{fig:main_results} (Qwen3-1.7B). To avoid selecting favorable examples, each prompt is the first of the seeded test draw used by the benchmarks. The strip at the top of each figure spans the whole prompt: the context in light grey, each kept token in red, and the non-prunable tokens $\mathcal{N}$ of Section~\ref{sec:method} (instruction, question and answer template; the scaffold) in darker grey at either end. The box below shows the part of the context under the bracket, chosen as the window of that length holding the most kept tokens. For each prompt we show the smallest budget $K \in \{64, 128, 256, 512, 1024\}$ at which the greedy answer equals that of the same model with the full context, or $K = 256$ where no budget does. The shown budgets keep between $0.4\%$ (NIAH) and $13\%$ (MultiNews) of the context.

\heading{Selection is global, not local.} In every sample the kept tokens fall in all ten deciles of the context, so the gate does not reduce to a recency window or to the start of the document. At the same time it concentrates the budget where the task needs it: on NIAH, $21$ of the $24$ tokens of the needle sentence, which sits at $78\%$ depth in a $15{,}843$-token haystack, are among the $64$ kept tokens, and the model answers exactly. The same holds for all eight NIAH prompts we checked, whose answer at $K = 64$ matches the reference.

\heading{What is kept, and what it costs.} Table~\ref{tab:kept_tokens} summarizes the five examples. On 2WikiMultihopQA, BioDEX and NIAH the gate keeps content: entities and years, the terms that describe a medical case, and, where the reference answer occurs verbatim in the context, a large share of its tokens (all of the needle's answer on NIAH). On RepoBench-P the selection is instead dominated by code structure: most kept tokens contain no letter, and they are spread thinly over the retrieved files, whereas on NIAH most kept tokens have a kept neighbor and form contiguous spans. This suggests that for code completion the post-gate layers use the kept tokens more as anchors on the structure of the retrieved files than as a copy of their content. The examples also show where pruning costs accuracy: on MultiNews the gate keeps mostly function words and under-keeps capitalized tokens, and the summary at $K = 256$ reports the right events but gets the names and places wrong, and for drug reaction extraction on BioDEX the pruned model recovers one of the two reactions found with the full context.

\begin{table}[tbp]
\centering
\small
\setlength{\tabcolsep}{3pt}
\begin{tabularx}{\linewidth}{@{}l >{\raggedright\arraybackslash}X >{\raggedright\arraybackslash}X@{}}
\toprule
Dataset & What is kept & Failure mode \\
\midrule
    RepoBench-P  &  code syntax rather than identifiers: commas and empty argument lists ``()'' dominate, line breaks are avoided, and kept tokens are isolated rather than spans  &                                                                                              wrong line, as with the full context \\
2WikiMultihopQA  &      entities and dates: names and years are over-kept and function words avoided; on the shown prompt, the answer phrase (``Film Award for \ldots Director'')  &  a different award at small budgets, exact at the shown one; multi-hop answers need the largest budgets to match the full context \\
      MultiNews  &                                                  mostly function words and scattered event words (``charged'', ``president''); names and places are under-kept  &                                          right events, but wrong name, occupation and place, consistent with the under-kept names \\
         BioDEX  &               terms of the diagnosis and outcome (``septic shock'', ``failure'') and the connectives of lists (``and'', ``or''); numbers are almost never kept  &                                                      misses reactions, but so does the full-context model, which bounds the score \\
       NIAH 16k  &                                                    the needle sentence as one contiguous span, plus rare subwords elsewhere (``-haired'', ``Lisp'', ``.html'')  &                                                                                                       none: exact at every budget \\
\bottomrule
\end{tabularx}

\caption{Summary of what \name keeps and how its answers fail. Examples refer to the prompts of Figures~\ref{fig:kept_repobench}--\ref{fig:kept_niah} at their shown budgets; the patterns hold over the first eight prompts of each test draw.}
\label{tab:kept_tokens}
\end{table}

\end{document}